\documentclass[conference,a4paper]{IEEEtran}
\IEEEoverridecommandlockouts

\usepackage[hidelinks]{hyperref}
\usepackage[cmex10]{amsmath}
\usepackage{amssymb,amsfonts}
\usepackage{dblfloatfix}

\usepackage[ruled,vlined]{algorithm2e}
\usepackage{graphicx}
\graphicspath{{Figures/PDF/}{Figures/PNG/}}

\usepackage{booktabs}
\usepackage{siunitx}
\usepackage[numbers,compress]{natbib}
\usepackage{texnames}
\usepackage{bm,bbm}
\usepackage{orcidlink}
\usepackage{cleveref}

\begin{document}

\title{
Tree Canopy Segmentation in Low-Data Regimes Using Pretrained Deep Models
}

\author{
\IEEEauthorblockN{David Szczecina}
\IEEEauthorblockA{
\textit{University of Waterloo}\\
Waterloo, Canada\\
david.szczecina@uwaterloo.ca}\\ 
\IEEEauthorblockN{Niloofar Azad}
\IEEEauthorblockA{
\textit{University of Waterloo}\\
Waterloo, Canada\\
n2azad@uwaterloo.ca}
\and
\IEEEauthorblockN{Hudson Sun}
\IEEEauthorblockA{
\textit{University of Waterloo}\\
Waterloo, Canada\\
hudson.sun@uwaterloo.ca}\\ 
\IEEEauthorblockN{Kyle Gao}
\IEEEauthorblockA{
\textit{University of Waterloo}\\
Waterloo, Canada\\
y56gao@uwaterloo.ca}
\and
\IEEEauthorblockN{Anthony Bertnyk}
\IEEEauthorblockA{
\textit{University of Waterloo}\\
Waterloo, Canada\\
abertnyk@uwaterloo.ca}\\
\IEEEauthorblockN{Lincoln Linlin Xu}
\IEEEauthorblockA{
\textit{University of Calgary}\\
Calgary, Canada\\
lincoln.xu@ucalgary.ca}

\thanks{Copyright 2026 IEEE. Published in the 2026 IEEE International Geoscience and Remote Sensing Symposium (IGARSS 2026)}
}

\maketitle
\begin{abstract}

Tree canopy detection from aerial imagery is an important task for environmental monitoring, urban planning, and ecosystem analysis. Simulating real-life data annotation scarcity, the Solafune Tree Canopy Detection competition \cite{Solafune_TreeCanopyDetection2025} provides a small and imbalanced dataset of only 150 annotated images, posing significant challenges for training deep models without severe overfitting. In this work, we evaluate five representative architectures, YOLOv11, Mask R-CNN, DeepLabv3, Swin-UNet, and DINOv2, to assess their suitability for canopy segmentation under extreme data scarcity. Our experiments show that pretrained convolution-based models, particularly YOLOv11 and Mask R-CNN, generalize significantly better than pretrained transformer-based models. DeeplabV3, Swin-UNet and DINOv2 underperform likely due to differences between semantic and instance segmentation tasks, the high data requirements of Vision Transformers, and the lack of strong inductive biases. These findings confirm that transformer-based architectures struggle in low-data regimes without substantial pretraining or augmentation and that differences between semantic and instance segmentation further affect model performance. We provide a detailed analysis of training strategies, augmentation policies, and model behavior under the small-data constraint and demonstrate that lightweight CNN-based methods remain the most reliable for canopy detection on limited imagery.

\end{abstract}

\begin{IEEEkeywords}
	\textbf{Deep Learning, Computer Vision, Object Segmentation, Remote Sensing, Forestry, Tree Canopy}.
\end{IEEEkeywords}

\section{Introduction}

Tree canopy detection from aerial or satellite imagery plays a critical role in a wide range of environmental applications, including biomass estimation, carbon accounting, biodiversity monitoring, and urban forestry. Accurate canopy mapping enables more effective policy decisions and more scalable ecological assessment. Despite its importance, high-quality canopy annotations are expensive to acquire, and many real-world datasets consist of only a few hundred samples. This creates a challenging setting for modern deep learning models, which typically rely on large, diverse datasets to achieve robust performance. The urgency of this work grows as wildfire seasons intensify across the globe and place unprecedented pressure on landscape monitoring, and the sparsity of labelled data remains a major challenge in AI-based wildfire research.

In this paper, we investigate five modern deep learning approaches, YOLOv11, Mask R-CNN, DeepLabv3, Swin-UNet, and DINOv2, to evaluate their ability to localize and segment tree canopies. We focus on understanding why some architectures succeed while others fail in limited-data settings, highlighting the importance of inductive bias, pretraining, and model capacity when training from only 150 images. Our findings show that convolution-based models remain considerably more robust than transformer-based models without large-scale pretraining.

\section{Background}
Remote sensing imagery presents challenges beyond standard vision tasks, including high intra-class variability from differences in resolution, atmospheric conditions, shadows, and land cover. Data from satellites, aircraft, drones, or ground platforms vary in sensor modality, ground sample distance (GSD), focal length, and viewing angle, causing canopy features to appear very differently across datasets. This scale variation demands models robust to changes in feature size, texture, and resolution, while limited annotated data requires strong inductive biases to prevent overfitting.

\paragraph{Convolution-based Approaches}
Convolutional Neural Networks (CNNs) such as U-Net, Mask-RCNN, DeepLabv3, and the YOLO family introduce useful spatial inductive biases such as locality and translation equivariance, making them naturally well-suited for segmentation when training data is scarce. Models like DeepLabv3 leverage dilated convolutions and multi-scale context through atrous spatial pyramid pooling (ASPP), enabling strong performance even with limited examples. Similarly, YOLOv11 offers a fast, anchor-free object detector with strong generalization due to extensive pretraining on large, diverse datasets. These properties help CNN-based models avoid severe overfitting on the 150-image Solafune dataset.

\paragraph{Transformer-based Approaches}
Vision Transformers (ViTs), including Swin-UNet and DINOv2, offer high modeling capacity and global receptive fields through self-attention mechanisms. However, ViTs lack the spatial priors inherent to CNNs and therefore require substantially more training data to learn stable visual representations. 
Swin-UNet introduces hierarchical attention and U-Net-like skip connections, but the core transformer blocks still remain highly data-dependent. 
DINOv2 provides powerful pretrained representations, yet fine-tuning with only 150 images leads to rapid overfitting and degraded segmentation quality. These limitations highlight the difficulty of applying transformer architectures to specialized remote sensing tasks without large-scale domain-relevant pretraining or heavy augmentation.

\paragraph{Tree Segmentation}
Traditional tree segmentation approaches rely heavily on LiDAR-derived 3D point cloud data and are commonly divided into two categories. The first class consists of methods that convert the point cloud into a canopy height model (CHM) and then apply surface-based analysis to detect individual tree crowns, as demonstrated in \citet{HUI2022103028} and \citet{ROUSSEL2020112061}. The second class comprises full 3D methods that operate directly on the raw point cloud. Among these, the point-based clustering algorithm, as proposed in \citet{8854321} and \citet{Pang03102021}, is a well-known and widely used method in this category due to its conceptual simplicity and its ability to capture tree-level structure. 

The advent of deep learning has substantially expanded the methodological landscape for tree segmentation. For LiDAR-based workflows, \citet{10.3389/fpls.2022.914974} applies YOLOv4 and R-CNN architectures to perform crown segmentation on height maps derived from airborne LiDAR, while \citet{f10090793} uses R-CNN to identify and segment tree trunks from 3D point cloud representations. Beyond LiDAR, a growing body of work has explored segmentation directly from aerial imagery, often reporting improved scalability and generalization. High-resolution aerial data, such as RGB or SAR imagery, are particularly attractive because they are easier to acquire at scale and typically provide higher spatial resolution than LiDAR. For example, \citet{VELASQUEZCAMACHO2023102025} employs YOLOv5 and Fast R-CNN for tree segmentation on high-resolution aerial and satellite images, and \citet{tolan2024canopy_dino} introduces a self-supervised vision-transformer approach using DINOv2 to generate large-scale tree height maps from RGB imagery.

Although these methods demonstrate strong segmentation accuracy, most require extensive training datasets to achieve their reported performance. In contrast, there are works deployed on small datasets. For example, \citet{TakahashiCNNViT} highlights cases of ViT models outperforming CNNs in medical image analysis tasks with fewer required training images. \citet{SAFONOVA2023103569} discusses approaches for addressing small datasets in remote sensing, but does not investigate transformer-based architectures. However, to the best of our knowledge, no existing work explicitly investigates the suitability of modern deep learning architectures, including ViTs and CNNs, for remote sensing scenarios where only a small training dataset is available. This motivates our study, in which we evaluate which contemporary architectures are most effective under limited-data conditions.
\section{Method}
In this study, we compare the performance of five architectures: YOLOv11 Seg, Mask R-CNN, DeepLabV3, Swin-UNet, and DINOv2. We randomly split the original training dataset into a training set and a validation set with a 4:1 ratio. For most of these models, we load a pretrained set of weights, which we then fine-tune on the sparse tree canopy dataset. For DINOv2, we freeze the pretrained backbone and design a dense pixel-based segmentation head.  

\begin{figure*}[!t]
    \centering
    \includegraphics[width=\linewidth]{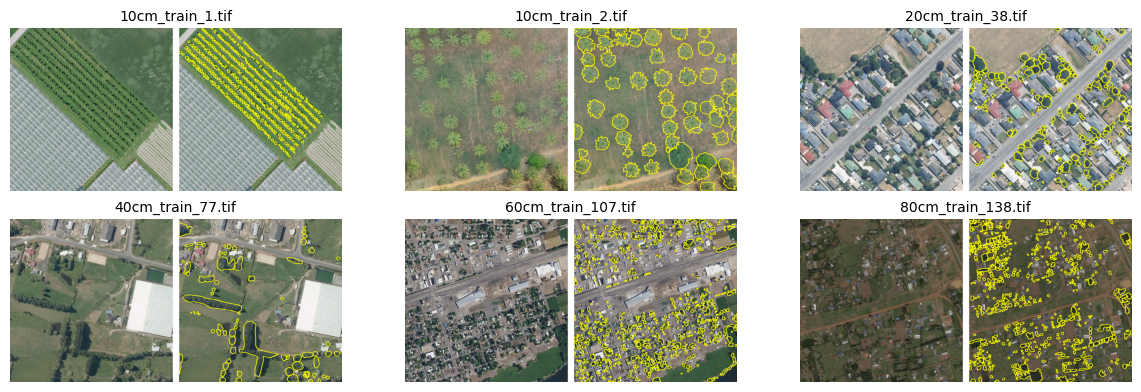}
    \caption{Examples of Training Images with Different Resolutions and Scenes and their Segmentation Labels.}
    \label{fig:sam_seg}
\end{figure*}

The Solafune Tree Canopy Detection dataset\cite{Solafune_TreeCanopyDetection2025} contains a training set and an evaluation (test) set, each with 150 aerial images in RGB tif format. Both datasets comprise five resolution groups - 10cm (38 images), 20cm (37 images), 40cm (25 images), 60cm (25 images), and 80cm (25 images). These images represent diverse land-cover scenes, including urban, rural, agricultural, industrial environments, and open fields. All image sizes are fixed to 1024x1024 pixels regardless of pixel resolution. Samples of raw images and segmentation from the training dataset, with varying resolutions and scenes, are displayed in \cref{fig:sam_seg}. The mean average precision (mAP) metric, designed for instance segmentation, is used as the evaluation metric for the Solafune Tree Canopy Detection dataset. We report the mAP on the Solafune competition's hidden test set. We also report the pixel-based accuracy on the validation set.

\textbf{YOLO}: All YOLOv11 \cite{yolo11_ultralytics} segmentation variants from nano through large were initialized with COCO pretrained weights and fine-tuned on the custom 1024×1024 canopy segmentation dataset without resizing.

\textbf{Mask R-CNN}: Mask R-CNN \cite{he2018maskrcnn} is initialized from its COCO pretrained weights and fine-tuned on the canopy dataset with random cropping and flipping while input images are resized to 640 by 640 to align with the original training scale and reduce compute.

\textbf{DeepLab}: DeepLabV3 \cite{chen2017rethinkingatrousconvolutionsemantic} is initialized from ImageNet-pretrained weights at 224 by 224 resolution, adapted to three canopy classes, fine-tuned on resized and extensively augmented inputs, and converted from semantic masks to polygonal instances for evaluation.

\textbf{DINO}: We adapt DINOv2 \cite{oquab2024dinov2learningrobustvisual} (specifically, its DINOv2 pretrained VIT backbone) for dense prediction by adding a trainable 1×1 convolutional segmentation head with bilinear upsampling while keeping the 86M-parameter backbone frozen.

\textbf{Swin-UNeT}: We employ Swin-UNet-Tiny \cite{cao2021swinunet} with ImageNet-pretrained weights and fine-tune the full model on the sparse canopy dataset, using its hierarchical encoder and patch-expanding decoder with skip connections to capture local and global context for dense prediction.

\vspace{2mm}
\section{Results}

The weighted mAP scores obtained on the Solafune Tree Canopy Detection hidden test set (obtained via competition upload) and the pixel accuracies on the validation set from the five models are reported in \Cref{tab:test_wmap}. As shown, all five models achieve close and high pixel-level accuracies for semantic segmentation, but YOLOv11 and Mask R-CNN substantially outperform the other three architectures, DeepLabv3, Swin UNet and DINOv2, in terms of weighted mAP scores, demonstrating a clear advantage in instance segmentation accuracy. 

 The detailed performance of each model is as follows:

\begin{table}[t]
\centering
\caption{Quantitative evaluation of tree canopy segmentation models on the Solafune dataset using validation pixel accuracy and test weighted mAP (IoU-based).} \label{tab:test_wmap}
\begin{tabular}{lcc}
\toprule
\textbf{Model} & \textbf{Val Pixel Accuracy} & \textbf{Test mAP} \\
\midrule
YOLOv11 Seg Large  & 0.86  & 0.281 \\
Mask R-CNN &  0.82  & 0.219 \\
DeepLabv3 &  0.82  & 0.038 \\
Swin-UNet     &     0.85     & 0.022 \\
DINOv2 &  0.82  & 0.021 \\
\bottomrule
\end{tabular}
\end{table}
\textbf{YOLO:} Performance improved with model capacity, with the large variant achieving the highest test mAP of 0.281, while smaller models converged faster but scored lower. Validation mAP underestimated generalization due to the small validation set, highlighting that larger models better capture the complexity of the task. Results are shown in \Cref{tab:yolo_wmap}. 

\textbf{Mask R-CNN:} Training was stable and validation performance plateaued early, yet the model achieved a reasonable test mAP of 0.22. This indicates good generalization despite limited validation gains.  

\textbf{DeeplabV3:} The model achieved strong validation metrics (pixel accuracy 0.82, mIoU 0.71) after extended training, showing effective learning. However, test performance was very low (weighted mAP 0.038) due to poor object-level mask conversion.  

\textbf{DINOv2:} Training on a small dataset was stable with improving pixel accuracy, reaching 0.82 on the validation set, but test mAP remained extremely low. Sparse and incomplete object-level mask predictions indicate that transformers require more data or domain-specific pretraining for small heterogeneous datasets.  

\textbf{Swin-UNet:} Pixel accuracy improved steadily on validation (0.85), demonstrating effective learning of spatial features. Test performance remained very low, highlighting limitations on small datasets.

\begin{figure*}
    \centering
    \includegraphics[width=1\linewidth]{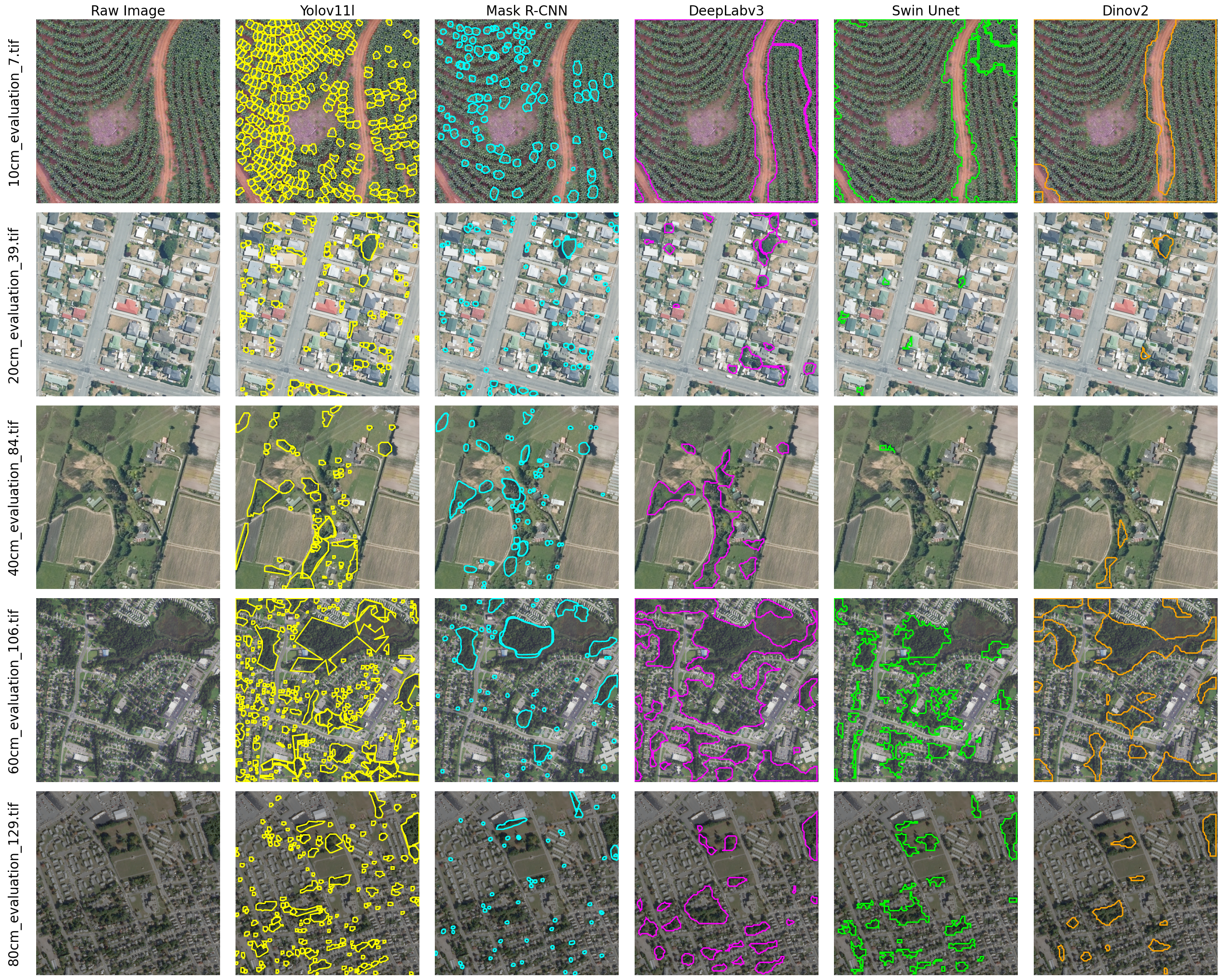}
    \caption{Qualitative Tree canopy Segmentation Results. Left to Right: Raw image, YOLOv11, Mask-RCNN, DeepLabv3, Swin-UNet, DinoV2}
    \label{fig:qualitative}
\end{figure*}

Qualitative results shown in \Cref{fig:qualitative} confirm the trend in \Cref{tab:test_wmap}, showing CNN-based architectures being less sensitive to region-based false negatives.

\begin{table}[t]
\centering
\caption{Quantitative comparison of YOLO-based models for tree canopy segmentation using validation pixel accuracy and test weighted mAP (IoU-based).} \label{tab:yolo_wmap}
\begin{tabular}{lcc}
\toprule
\textbf{Model} & \textbf{Val mAP} & \textbf{Test mAP} \\
\midrule
YOLOv11 Seg Large  & 0.189  & 0.281 \\
YOLOv11 Seg Medium  & 0.187  & 0.279 \\
YOLOv11 Seg Small  &  0.177 & 0.257 \\
YOLOv11 Seg Nano  &  0.164 & 0.249 \\
\bottomrule
\end{tabular}
\end{table}

\section{Discussion}

We fine-tuned models from several influential object detection and instance segmentation models. We found that transformer-based architectures often failed to produce reliable bounding boxes and behaved more like pixel-level semantic segmentation models. This failure produced low instance-level metrics such as weighted mAP while leaving pixel-level accuracy relatively high. The same pixel-centric behavior appears in DeepLabV3. Comparing architectures makes the disparity clear. Mask R-CNN, designed for instance segmentation, substantially outperformed DeepLabV3, which was designed for semantic segmentation, on weighted mAP. Semantic segmentation models therefore do not generalize well to instance-level tasks without instance-based prediction branches. 

Training curves provide further evidence. After 100 epochs DeepLabV3 reached higher pixel accuracy while DINOv2 began to overfit, which suggests modern ViTs can learn less effectively than older CNNs on small datasets. A related caveat is that pretrained object detectors produce object-level outputs by different mechanisms, and fine-tuning the backbone together with an instance-prediction head may interact unexpectedly with bounding-box creation in DeepLabV3, Swin-UNet, and DINOv2, resulting in their behavior approaching semantic segmentation models while ignoring instance-based information. 

Future work should explore hybrid CNN–ViT designs and more sophisticated fine-tuning strategies tailored to small-scale remote sensing datasets.
\section{Conclusion}
Tree canopy detection is vital for environmental monitoring but hampered by scarce annotations. Our fine-tuning experiments on the 150-image Solafune dataset found that convolutional models YOLOv11 and Mask R-CNN were the most stable and accurate. Their spatial inductive biases and large-scale pretraining allowed them to generalize despite limited supervision. These results reinforce the introduction’s point that dependable canopy mapping for applications such as carbon accounting and wildfire monitoring depends on architectures or priors that compensate for small, costly datasets. Overall, the results indicate that in extremely low-data remote-sensing settings, lightweight CNN architectures and instance-segmentation frameworks remain considerably more robust than pretrained ViT-based object detector design.


\small
\clearpage
\bibliographystyle{IEEEtranN}
\bibliography{references}

@article{tolan2024canopy_dino,
title = {Very high resolution canopy height maps from RGB imagery using self-supervised vision transformer and convolutional decoder trained on aerial lidar},
journal = {Remote Sensing of Environment},
volume = {300},
pages = {113888},
year = {2024},
issn = {0034-4257},
doi = {https://doi.org/10.1016/j.rse.2023.113888},
url = {https://www.sciencedirect.com/science/article/pii/S003442572300439X},
author = {Jamie Tolan and Hung-I Yang and Benjamin Nosarzewski and Guillaume Couairon and Huy V. Vo and John Brandt and Justine Spore and Sayantan Majumdar and Daniel Haziza and Janaki Vamaraju and Theo Moutakanni and Piotr Bojanowski and Tracy Johns and Brian White and Tobias Tiecke and Camille Couprie},
}

@misc{Solafune_TreeCanopyDetection2025,
  author       = {{Solafune, Inc.}},
  title        = {Tree Canopy Detection — Competition Overview},
  howpublished = {\url{https://solafune.com/competitions/26ff758c-7422-4cd1-bfe0-daecfc40db70?menu=about&tab=overview}},
  note         = {Accessed: 2025-10-09},
  year         = {2025}
}

@article{cao2021swinunet,
  title={Swin-Unet: Unet-like pure Transformer for medical image segmentation},
  author={Cao, Hu and Wang, Yue and Chen, Jiarui and Jiang, Dongsheng and Zhang, Xin and Tian, Qi and Wang, Yunhai},
  journal={arXiv preprint arXiv:2105.05537},
  year={2021}
}

@article{HUI2022103028,
title = {Multi-level self-adaptive individual tree detection for coniferous forest using airborne LiDAR},
journal = {International Journal of Applied Earth Observation and Geoinformation},
volume = {114},
pages = {103028},
year = {2022},
issn = {1569-8432},
doi = {https://doi.org/10.1016/j.jag.2022.103028},
url = {https://www.sciencedirect.com/science/article/pii/S1569843222002163},
author = {Zhenyang Hui and Penggen Cheng and Bisheng Yang and Guoqing Zhou},
}

@article{ROUSSEL2020112061,
title = {lidR: An R package for analysis of Airborne Laser Scanning (ALS) data},
journal = {Remote Sensing of Environment},
volume = {251},
pages = {112061},
year = {2020},
issn = {0034-4257},
doi = {https://doi.org/10.1016/j.rse.2020.112061},
url = {https://www.sciencedirect.com/science/article/pii/S0034425720304314},
author = {Jean-Romain Roussel and David Auty and Nicholas C. Coops and Piotr Tompalski and Tristan R.H. Goodbody and Andrew Sánchez Meador and Jean-François Bourdon and Florian {de Boissieu} and Alexis Achim},
}

@ARTICLE{8854321,
  author={Williams, Jonathan and Schönlieb, Carola-Bibiane and Swinfield, Tom and Lee, Juheon and Cai, Xiaohao and Qie, Lan and Coomes, David A.},
  journal={IEEE Transactions on Geoscience and Remote Sensing}, 
  title={3D Segmentation of Trees Through a Flexible Multiclass Graph Cut Algorithm}, 
  year={2020},
  volume={58},
  number={2},
  pages={754-776},
  doi={10.1109/TGRS.2019.2940146}}

@article{Pang03102021,
author = {Yong Pang and Weiwei Wang and Liming Du and Zhongjun Zhang and Xiaojun Liang and Yongning Li and Zuyuan Wang},
title = {Nyström-based spectral clustering using airborne LiDAR point cloud data for individual tree segmentation},
journal = {International Journal of Digital Earth},
volume = {14},
number = {10},
pages = {1452--1476},
year = {2021},
publisher = {Taylor \& Francis},
doi = {10.1080/17538947.2021.1943018},
URL = { 
https://doi.org/10.1080/17538947.2021.1943018
},
eprint = { 
https://doi.org/10.1080/17538947.2021.1943018
}

}

@ARTICLE{10.3389/fpls.2022.914974,
  
AUTHOR={Sun, Chenxin  and Huang, Chengwei  and Zhang, Huaiqing  and Chen, Bangqian  and An, Feng  and Wang, Liwen  and Yun, Ting },
         
TITLE={Individual Tree Crown Segmentation and Crown Width Extraction From a Heightmap Derived From Aerial Laser Scanning Data Using a Deep Learning Framework},
        
JOURNAL={Frontiers in Plant Science},
        
VOLUME={Volume 13 - 2022},

YEAR={2022},

URL={https://www.frontiersin.org/journals/plant-science/articles/10.3389/fpls.2022.914974},

DOI={10.3389/fpls.2022.914974},

ISSN={1664-462X},
}

@Article{f10090793,
AUTHOR = {Wang, Jiamin and Chen, Xinxin and Cao, Lin and An, Feng and Chen, Bangqian and Xue, Lianfeng and Yun, Ting},
TITLE = {Individual Rubber Tree Segmentation Based on Ground-Based LiDAR Data and Faster R-CNN of Deep Learning},
JOURNAL = {Forests},
VOLUME = {10},
YEAR = {2019},
NUMBER = {9},
ARTICLE-NUMBER = {793},
URL = {https://www.mdpi.com/1999-4907/10/9/793},
ISSN = {1999-4907},

}

@article{VELASQUEZCAMACHO2023102025,
title = {Implementing Deep Learning algorithms for urban tree detection and geolocation with high-resolution aerial, satellite, and ground-level images},
journal = {Computers, Environment and Urban Systems},
volume = {105},
pages = {102025},
year = {2023},
issn = {0198-9715},
doi = {https://doi.org/10.1016/j.compenvurbsys.2023.102025},
url = {https://www.sciencedirect.com/science/article/pii/S0198971523000881},
author = {Luisa Velasquez-Camacho and Maddi Etxegarai and Sergio de-Miguel},
}

@misc{he2018maskrcnn,
      title={Mask R-CNN}, 
      author={Kaiming He and Georgia Gkioxari and Piotr Dollár and Ross Girshick},
      year={2018},
      eprint={1703.06870},
      archivePrefix={arXiv},
      primaryClass={cs.CV},
      url={https://arxiv.org/abs/1703.06870}, 
}

@article{TakahashiCNNViT,
author = {Takahashi, Satoshi and Sakaguchi, Yusuke and Kouno, Nobuji and Takasawa, Ken and Ishizu, Kenichi and Akagi, Yu and Aoyama, Rina and Teraya, Naoki and Shinkai, Norio and Machino, Hidenori and Kobayashi, Kazuma and Asada, Ken and Komatsu, Masaaki and Kaneko, Syuzo and Sugiyama, Masashi and Hamamoto, Ryuji},
year = {2024},
month = {09},
pages = {84},
title = {Comparison of Vision Transformers and Convolutional Neural Networks in Medical Image Analysis: A Systematic Review},
volume = {48},
journal = {Journal of Medical Systems},
doi = {10.1007/s10916-024-02105-8}
}

@article{SAFONOVA2023103569,
title = {Ten deep learning techniques to address small data problems with remote sensing},
journal = {International Journal of Applied Earth Observation and Geoinformation},
volume = {125},
pages = {103569},
year = {2023},
issn = {1569-8432},
doi = {https://doi.org/10.1016/j.jag.2023.103569},
url = {https://www.sciencedirect.com/science/article/pii/S156984322300393X},
author = {Anastasiia Safonova and Gohar Ghazaryan and Stefan Stiller and Magdalena Main-Knorn and Claas Nendel and Masahiro Ryo},
}

@software{yolo11_ultralytics,
  author = {Glenn Jocher and Jing Qiu},
  title = {Ultralytics YOLO11},
  version = {11.0.0},
  year = {2024},
  url = {https://github.com/ultralytics/ultralytics},
  orcid = {0000-0001-5950-6979, 0000-0003-3783-7069},
  license = {AGPL-3.0}
}

@misc{chen2017rethinkingatrousconvolutionsemantic,
      title={Rethinking Atrous Convolution for Semantic Image Segmentation}, 
      author={Liang-Chieh Chen and George Papandreou and Florian Schroff and Hartwig Adam},
      year={2017},
      eprint={1706.05587},
      archivePrefix={arXiv},
      primaryClass={cs.CV},
      url={https://arxiv.org/abs/1706.05587}, 
}

@misc{oquab2024dinov2learningrobustvisual,
      title={DINOv2: Learning Robust Visual Features without Supervision}, 
      author={Maxime Oquab and Timothée Darcet and Théo Moutakanni and Huy Vo and Marc Szafraniec and Vasil Khalidov and Pierre Fernandez and Daniel Haziza and Francisco Massa and Alaaeldin El-Nouby and Mahmoud Assran and Nicolas Ballas and Wojciech Galuba and Russell Howes and Po-Yao Huang and Shang-Wen Li and Ishan Misra and Michael Rabbat and Vasu Sharma and Gabriel Synnaeve and Hu Xu and Hervé Jegou and Julien Mairal and Patrick Labatut and Armand Joulin and Piotr Bojanowski},
      year={2024},
      eprint={2304.07193},
      archivePrefix={arXiv},
      primaryClass={cs.CV},
      url={https://arxiv.org/abs/2304.07193}, 
}

\end{document}